\documentclass[pmlr,twocolumn,10pt]{jmlr} 

\usepackage{booktabs}
\usepackage{siunitx}

\newcommand{\equal}[1]{{\hypersetup{linkcolor=black}\thanks{#1}}}

\theorembodyfont{\upshape}
\theoremheaderfont{\scshape}
\theorempostheader{:}
\theoremsep{\newline}

\jmlrvolume{LEAVE UNSET}
\jmlryear{2022}
\jmlrsubmitted{LEAVE UNSET}
\jmlrpublished{LEAVE UNSET}
\jmlrworkshop{Conference on Health, Inference, and Learning (CHIL) 2022} 

\title[Semi-Markov Offline Reinforcement Learning for Healthcare]{Semi-Markov Offline Reinforcement Learning for Healthcare}

\author{%
\Name{Mehdi Fatemi}\equal{These authors contributed equally} \Email{mehdi.fatemi@microsoft.com}\\
\addr Microsoft Research
\AND
\Name{Mary Wu}\footnotemark[1] \Email{mary@cs.toronto.edu}\\
\addr University of Toronto
\AND
\Name{Jeremy Petch} \Email{petchj@hhsc.ca} \\
\addr Hamilton Health Sciences
\AND
\Name{Walter Nelson} \Email{nelsonwa@hhsc.ca}\\
\addr Hamilton Health Sciences
\AND
\Name{Stuart J. Connolly} \Email{stuart.connolly@phri.ca} \\
\addr Population Health Research Institute
\AND
\Name{Alexander Benz} \Email{alexander.benz@phri.ca}\\
\addr Population Health Research Institute
\AND
\Name{Anthony Carnicelli} 
\Email{anthony.carnicelli@duke.edu}\\
\addr Duke University
\AND
\Name{Marzyeh Ghassemi} 
\Email{mghassem@mit.edu}\\
\addr Massachusetts Institute of Technology
}
\begin{document}

\maketitle

\begin{abstract}
Reinforcement learning (RL) tasks are typically framed as Markov Decision Processes (MDPs), assuming that decisions are made at fixed time intervals. However, many applications of great importance, including healthcare, do not satisfy this assumption, yet they are commonly modelled as MDPs after an artificial reshaping of the data. In addition, most healthcare (and similar) problems are \emph{offline} by nature, allowing for only retrospective studies. To address both challenges, we begin by discussing the Semi-MDP (SMDP) framework, which formally handles actions of variable timings. We next present a formal way to apply SMDP modifications to nearly any given value-based offline RL method. We use this theory to introduce three SMDP-based offline RL algorithms, namely, SDQN, SDDQN, and SBCQ. We then experimentally demonstrate that only these SMDP-based algorithms learn the optimal policy in variable-time environments, whereas their MDP counterparts do not. Finally, we apply our new algorithms to a real-world offline dataset pertaining to \emph{warfarin dosing for stroke prevention} and demonstrate similar results.
\end{abstract}

\paragraph*{Data and Code Availability}
Access to the Randomized Control Trial data used for the warfarin study is governed by the COMBINE AF executive committee. Investigators interested in working with these data should contact a member of the committee to discuss potential collaboration \citep{Carnicelli2021}. The code is available at https://github.com/mary-wu/smdp.

\section{Introduction}
\label{sec:intro}
Many healthcare and medical problems are sequential decision-making tasks: the patient's health condition is monitored, medical interventions are administered accordingly, and the cycle repeats until a termination point. A formal way to model such problems is through Markov Decision Processes (MDPs) \citep{puterman_book}. In the MDP setting, patients are often modelled by their health \textit{state}. The state accounts for a sufficient history of the patient's health records, as well as demographic or other relevant information, which together make the state self-contained at any time point~\citep{ghassemi2014unfolding,ghassemi2015multivariate,wu2016ssam,ghassemi2017predicting,suresh2017clinical,raghu2017continuous}. Solving an MDP may be achieved through a wealth of algorithmic approaches, which are collectively known as \textit{reinforcement learning} (RL) \citep{bertsekas_neuro,sutton_book}. However, there are two significant challenges in many healthcare problems. First, observations are sporadic, which violates the MDP's required regular spacing of time intervals. Second, datasets are typically offline without an opportunity to interact. This paper will propose a class of algorithms to address both concerns.

A motivating example is warfarin dose management. Warfarin is an anticoagulant commonly prescribed to reduce the risk of stroke associated with atrial fibrillation \citep{warfarin_intro}. This example is of particular interest as stroke is the second leading cause of death worldwide \cite{who}, warfarin continues to be used around the globe, and clinicians struggle to prescribe it effectively because of its complex pharmacodynamics. Warfarin dosing for atrial fibrillation is easily modelled as a sequential decision-making task: patients visit the clinic at certain time intervals to monitor their health, during which a physician may adjust their dose. At each visit, the patient’s International Normalized Ratio (INR) is measured and used, along with the patient’s medical history and demographic information, to potentially modify the dosage. Although this is a well-defined decision-making process, we cannot directly model this as an MDP due to inconsistent times between clinical visits (see Section \ref{sec:warfarin_mdp}). 

In general terms, the MDP framework contains no notion of a course of action persisting over a variable period of time \citep{sutton-smdp}. More precisely, the standard MDP framework does not include temporal abstraction or temporally extended actions. Temporal abstraction can be introduced into RL in a variety of ways, see for example \cite{RING1991343, WIXSON1991368, Schmidhuber91neuralsequence, Singh1992, Chrisman1994ReasoningAP, Sutton1995modeling, Precup97macro, Precup97multitimemodels}. Among these methods, using \textit{options} \citep{Pericup1998options} has versatility, solid mathematical exposition, and similarity of the resultant theory to the standard RL framework. An option is a fixed policy with an initiation function, which signals where the policy can be selected, and a termination function, which specifies when the policy is terminated while it is running. While an option is active, the primitive actions are chosen according to its policy until the termination function becomes true, at which point the next option is chosen. Options have been utilized to formally introduce temporal abstractions to the MDP framework \citep{sutton-smdp} through the Semi-MDP (SMDP) framework \citep{Bradtke_undated-tf, Parr1998-au}, by replacing actions with options. This makes SMDPs appropriate for modeling unregulated discrete-event problems over continuous or discrete time. There are several other methods that are built on continuous-time dynamic programming (CDP). For example, Marked Temporal Processes \citep{Aalen2008, Zarezade2017} are a domain-specific application of CDP (they use classic HJB results, which are also core to SMDP theory \citep{Bradtke_undated-tf}). We however note that none of these methods are considered inherently different methodologies; hence, we do not discuss them in this paper nor compare our results against them. Importantly, we also remark that none of these classic or recent materials pertain to the offline case, which we discuss next.

Another important characteristic of many healthcare and treatment management problems is their \textit{offline} nature \citep{lange2012batch}, i.e., the dataset is static and must be analysed in a retrospective manner due to diverse safety, ethical, or legal constraints \citep{fatemi2021medical}. While offline RL literature has advanced substantially in recent years, proposed offline RL algorithms still target MDP scenarios \citep{lange2012batch, fujimoto2019off}. One core contribution of this paper is to introduce a systematic way to construct SMDP versions of common offline RL methods. This new class of algorithms is sound and easy to implement once the MDP version is given. Of particular interest, we introduce the SBCQ algorithm, an SMDP version of Batch-Constrained Q-learning (BCQ) \citep{fujimoto2019off}. SBCQ addresses the over-estimation issue of offline RL methods the same way that BCQ does, but extends the framework to handle Semi-Markov regimes with arbitrary options. 

While options are great tools to properly address the irregular timing of healthcare scenarios, the standard SMDP framework requires full observability of intra-option rewards. This is not normally the case in healthcare (and perhaps various other similar domains), since both the state and the reward are monitored solely at the time that the patient comes to the clinic. To address this issue, we choose to use linear interpolation and will argue how this solution is medically sound in the case of the stroke problem. In addition, we are interested in the case where options are given rather than learned. This is because in our healthcare setting (and similar problems), the options are used to model the maintenance of a given prescription for various time intervals.

The rest of the paper is organized as follows: Section 2 discusses important related work. Section 3 explains the setting and reviews important results from the literature, which sets the stage for our main technical contributions, presented in Section 4. We then present a simple domain in Section 5 to concretely demonstrate how our methods work in comparison to the MDP variations, and contrast SBCQ with simpler SMDP algorithms. Finally, in Section 6, we apply our methods to the warfarin dosage management problem and report similar conclusions in real-world settings.

\section{Related Work}

{\raggedright\textbf{RL for Health:}\setlength{\parindent}{15pt}}
RL has been the subject of much focus in health~\citep{yu2019reinforcement}, with particular emphasis on sepsis with various goals: seeking to develop optimal treatment recommendation policies~\citep{henry2015targeted,futoma2017improved,komorowski2018artificial,saria2018individualized,raghu2017continuous,peng2018improving,li2019optimizing,tang2020clinician}, treatment avoidance~\citep{fatemi2021medical}, and learning  set-valued  policies with the goal of human-in-the-loop treatment selection~\citep{shengpu2020}. All such prior work aggregate observations at equally distributed time points such that these processes can be modelled as MDPs. These are in direct contrast to our approach that allows arbitrary time stamps for observations. 

{\raggedright\textbf{SMDP in Deep RL:}\setlength{\parindent}{15pt}}
While SMDP theory has a long history, its deployment to Deep RL settings is fairly recent and limited. For example, \cite{Schmoll2020} uses SMDP in a DQN setting for the task of collecting stochastic, spatially distributed resources (stochastic resource collection), and \cite{Bellinger2020} broadly applies SMDP to DQN and Recurrent DQN. However, to the best of our knowledge, none of the few existing publications investigate offline scenarios, which is one of the core considerations of the current paper.

{\raggedright\textbf{Offline RL:}\setlength{\parindent}{15pt}} Offline RL has been studied for a long time in the context of batch RL \citep{lange2012batch}. Importantly, the performance of \textit{off-policy} RL algorithms \citep{sutton_book} degrade drastically in fully \textit{offline} or batch settings as they need to sufficiently explore the environment with additional interactions \citep{jaques2019way,fujimoto2019off}. The challenges are significantly intensified when the dataset is limited and exploration of various states and actions is not available ~\citep{bertsekas_neuro, kushner2003}. Overfitting to data-collection artifacts in offline cases is another significant issue \citep{francois2019overfitting, sinha2021s4rl, agarwal2020optimistic}. Moreover, estimation errors due to limited data may further lead to mistimed or inappropriate decisions with adverse safety consequences ~\citep{rebba2006validation}. For a more recent, yet partial, survey of MDP-based offline RL methods see \cite{Levine2020}. 
Since most healthcare and medical-related sequential decision-making/reasoning problems are offline, this paper studies the proposed SMDP algorithms in fully offline scenarios.

\section{Background}

\subsection{Markov Decision Processes}

We consider discrete-time episodic Markov Decision Processes (MDPs), where an agent interacts with the environment at each of the lowest-level discrete time steps $t=1, 2, \dots$. An MDP is defined as a tuple $(\mathcal{X}, \mathcal{A}, R, T, \gamma)$. $\mathcal{X}$ and $\mathcal{A}$ are respectively the state and action spaces, $T$ is the transition kernel $T(x,a,\cdot)=P(\cdot|x,a)$ that gets the probability of transitions if action $a$ is selected at state $x$, $\gamma \in [0,1]$ is the discount factor, and $R: \mathcal{X}\times\mathcal{A}\times\mathcal{X}\mapsto \mathbb{R}$ is the reward function. A stationary policy $\pi$ maps each state $x\in\mathcal{X}$ to a probability distribution over the action space $\mathcal{A}$. We define a terminal state as the final state at which the environment terminates (in healthcare, it corresponds to the last point of a patient's recorded trajectory). The set of all terminal states is denoted by $\mathcal{X}_{T}\subset \mathcal{X}$. Mathematically, a terminal state is absorbing (self-transition w.p. 1) with zero reward afterwards. All terminal states are zero-valued, but the transitions to them may be associated with a non-zero reward. 

On each time step $t$, the agent perceives the environment's state, $X_t = x$, and selects a primitive action, $A_t=a$, according to a policy $\pi(\cdot|x)$. The action influences the state and together with the internal dynamics of the environment, it transitions to a new state $X_{t+1}=x'$ at the next discrete time $t+1$. This transition may incur a reward of $R_{t+1}(x,a,x')$. We define \textit{return} as the sum of discounted rewards as the agent interacts with the environment. The value function corresponding to a policy $\pi$ is denoted by $Q^{\pi}(x,a)$ and defined as the expected return if the agent selects action $A_{t}=a$ at state $X_{t}=x$ and acts according to $\pi$ afterwards.

We define optimal value and state-value functions, respectively as $Q^*(x,a)=\max_{\pi}Q^{\pi}(x,a)$ and $V^*(x)=\max_{\pi}V^{\pi}(x)=\max_{a'}Q^{*}(x,a')$.

\subsection{Options and Semi Markov Decision Processes}

Semi-Markov Decision Processes (SMDPs) are a generalization of MDPs that handle actions with variable duration. More precisely, an SMDP is similar to an ordinary MDP, with the difference being that transitions may have stochastic time duration \citep{Parr1998-au}. The original SMDP theory treats the extended actions as black boxes over continuous time \citep{Bradtke_undated-tf, Parr1998-au}. To have a more versatile view, \cite{sutton-smdp} built on that and introduced a superimposed version, where each extended action is seen as an \textit{option}. We adopt their version in this paper and review the formalism in the rest of this section. 

An option is defined as the tuple $(\mathcal{I}, \pi, \beta)$, where $\pi: \mathcal{X}\times\mathcal{A}\mapsto [0,1]$ is a policy, $\mathcal{I}\subseteq \mathcal{X}$ is the initiation set where the policy can be chosen, and $\beta: \mathcal{X}\mapsto [0,1]$ is the termination function. If an option is active, actions are selected according to its policy $\pi$ until the option terminates stochastically according to $\beta$. For instance, the option of ``opening-the-door'' specifies the (possibly stochastic) sequence of all the primitive actions (e.g., all the muscle contractions) involved in reaching the door knob, turning it, pulling/pushing the door with proper force, and releasing the knob. More precisely, the sequence of such primitive actions can be seen as initiating a policy $\pi$ at some proper set of states $\mathcal{I}$ (e.g., where the knob is reachable), running, and then terminating it once the knob is released (where $\beta=1$).

Policies and termination functions are called Markov if they are only a function of the current state. On the other hand, an option's policy is called \textit{Semi-Markov} if it is a function of any event from the \textit{history} of the option. If an option is selected at time $t$, we define the history from $t$ to any time $t'>t$ by $h_{tt'}=\{ s_{t}, a_{t}, r_{t+1}, s_{t+1}, \dots, s_{t'} \}$. Remark that Semi-Markov policies are less unconstrained than non-stationary policies in that Semi-Markovness only allows being function of events back to time $t$, but not before that. Similarly, we can define the termination function $\beta$ to be Semi-Markov if it is a function of the option's history. For an option $o=(\mathcal{I}, \pi, \beta)$, we say $o$ is Markov if both $\pi$ and $\beta$ are Markov, and we say $o$ is Semi-Markov if any of $\pi$ or $\beta$ is Semi-Markov. 

The concept of policy can be applied to options as well. Formally, if $\mathcal{O}$ is a set of options defined on an MDP $M=(\mathcal{X}, \mathcal{A}, R, T, \gamma)$, a policy $\mu(o|x)$ is defined as the probability of selecting $o=(\mathcal{I}, \pi, \beta)\in \mathcal{O}$ at state $x\in\mathcal{X}$. If $x\not\in\mathcal{I}$, then $\mu(o|x)=0$. Hence, a policy over a set of options induces a \textit{flat policy} at the primitive actions level by selecting an option, waiting until it terminates, selecting the next option and so on. The flat policy is defined as the concatenation of all the option-level policies. Note that the flat policy is unlikely to be Markov even if both $\mu$ and all the options are Markov. We next extend the concept of value functions to \textit{Semi-Markov flat policies} as 
\begin{align}
    \nonumber V^{\pi}(x) \doteq \mathbb{E}\left[\sum_{j=0}^{\infty} \gamma^{j}R_{t+j+1}|\mathcal{E}(\pi,x,t)\right]
\end{align}
where $\mathcal{E}(\pi,x,t)$ denotes the event of $\pi$ being initiated at state $x$ at time $t$. By definition, the value of a state under an option-level policy can then be defined as the value of the state under the corresponding flat policy: $V^{\mu}(x)\doteq V^{\textsf{flat}(\mu)}(x)$ for all $x\in\mathcal{X}$. We also define $Q^{\mu}(x,o)$ as the value of selecting option $o=(\mathcal{I}, \pi, \beta)\in\mathcal{O}$ at state $x\in\mathcal{I}$ under policy $\mu$; formally,
\begin{align}
    \nonumber Q^{\mu}(x,o) \doteq \mathbb{E}\left[\sum_{j=0}^{\infty} \gamma^{j}R_{t+j+1}|\mathcal{E}(o\mu,x,t)\right]
\end{align}
where $o\mu$ denotes the Semi-Markov policy (composing $o$ and $\mu$) that first follows the policy of $o$ until it terminates and then starts choosing another option according to $\mu$ in the resultant state and follows its policy, and so on. 

SMDPs \citep{Bradtke_undated-tf} are defined in a similar way as the MDPs, but with temporally extended actions, the selection of which depends on the history from each given state. 
Theorem 1 of \citep{sutton-smdp} states that any fixed set of options defined on an MDP establishes a discrete time SMDP. That is, the base problem is an MDP, over which extended actions are shaped. This view provides flexibility in terms of what happens inside each option, which can be examined, altered, learned, and planned.
 
We are interested in the case where a fixed set of options is given and we would like to determine the best strategy for selecting among the options when the current option terminates. Let us denote such a set of options as $\mathcal{O}$. Remark that if $\mathcal{O}$ does not contain all the primitive actions and/or any of the options lasts for more than one step, then achieving the optimal values at the level of primitive actions may \textit{not} be feasible. Still, it is important to achieve the best results given the options. Of note, having more flexibility in terms of option design can improve the ultimate performance. For example, it can be proved that having the choice to terminate the current option whenever the agent wants (called option interruption) directly improves the values of all states. This extension can easily be applied using our pipeline. However, at this stage we assume that no observation is available between subsequent visits; hence, option interruption is also infeasible. 

Let $\mu: \mathcal{X}\times\mathcal{O}\mapsto [0,1]$ be an option-level policy that selects options from $\mathcal{O}$ at states where the options' initiation sets allow, and let $\Pi(\mathcal{O})$ denote the set of all such policies. Expanding the optimal value function at the option level, it then follows:
\begin{align}
    \nonumber V_{\mathcal{O}}^{*}(x) &\doteq \max_{\mu\in\Pi(\mathcal{O})} V^{\mu} \\
    \nonumber &= \max_{o\in\mathcal{O}}\mathbb{E}[R_{t+1}+\dots+\gamma^{k-1}R_{t+k} \\
    \nonumber &\quad\quad\quad\quad\quad +\gamma^{k}V_{\mathcal{O}}^{*}(x_{t+k})~|~\mathcal{E}(o,x,t)] \\
    &= \max_{o\in\mathcal{O}}\mathbb{E}[\rho(x,o) + \gamma^{k}V_{\mathcal{O}}^{*}(x')~|~\mathcal{E}(o,x)] \label{eq:Vo}
\end{align}
where $k$ is the duration of $o$ when taken at $x$, $x'$ is the state where the option terminates, and $\rho(s,o)$ is the \textit{discounted cumulative rewards} in the course of option's operation. We also dropped $t$ for clarity. Similarly, $Q_{\mathcal{O}}^{*}$ satisfies: 
\begin{align}
    Q_{\mathcal{O}}^{*}(x,o) = \mathbb{E}[\rho(x,o) + \gamma^{k}\max_{o'\in\mathcal{O}}Q_{\mathcal{O}}^{*}(x',o')~|~\mathcal{E}(o,x)] \label{eq:Qo}
\end{align}
Importantly, remark that these Bellman equations incur a factor of $\gamma^{k}$ for the bootstrapping term. This is to be expected since $x'$ is $k$ steps away from $x$. Nevertheless, this multiplier, along with the fact that $\rho(x,o)$ is the discounted return rather than the immediate reward, directly impact learning the optimal values. Equations (\ref{eq:Vo}) and (\ref{eq:Qo}) can directly be used in planning algorithms such as (SMDP-) value- or policy-iteration (see \cite{sutton-smdp} for details and examples). Similarly, for TD-like learning algorithms, the update rule must be modified accordingly. Most prominently, the \textit{SMDP Q-learning} update \citep{Bradtke_undated-tf, sutton-smdp} takes the following form:
\begin{align} \label{eq:smdp:q-learning}
    \nonumber Q(x,o)\longleftarrow (&1-\alpha) Q(x,o) \\
    &+ \alpha \left[\rho(x,o) +\gamma^{k}\max_{o'\in \mathcal{O}}Q(x',o')\right]
\end{align}
where $\alpha$ is the learning rate. The SQ-learning algorithm of \eqref{eq:smdp:q-learning} is guaranteed to converge to the $Q^*_{\mathcal{O}}$ under similar conditions to the standard Q-learning \citep{Watkins89-phd, Watkins1992-ty}. 

The core technical contribution of our work is to apply similar modifications to the update rules of state-of-the-art offline RL algorithms to yield their SMDP counterparts.

\newcommand{\wD}{\widetilde{\mathcal{D}}}
\SetKwRepeat{Repeat}{repeat}{until}%
\begin{algorithm2e*}[t!]
\caption{Generic algorithm for SMDP-based offline RL.}
\label{alg:smdp}
\DontPrintSemicolon
\LinesNumbered
\SetNoFillComment
\KwData{$\mathcal{D}'$}
\KwIn{randomly initialized $Q_{\theta}$, $\gamma$}
\KwOut{learned $Q_{\theta}$}
$j\longleftarrow 0\quad$ \tcc{$j:=$ total iteration counter}
\For{epoch $=1,M$}{
$\wD\longleftarrow \mathcal{D}'$\;
\Repeat{$\wD$ is empty}{
sample minibatch $\{(x_i,o_i,\rho_i,x'_i,k_i)\} \sim U(\wD)$  \quad \tcc{$i :=$ index inside the minibatch}
\For{all $i$}{
$y_i\longleftarrow \left\{
    \begin{array}{ll}
        \rho_i & \mbox{if } x'_i \mbox{ is terminal} \\
        \rho_i + \gamma^{k_i} \max_{o'_i} Q_{\theta_j} (x'_i,o'_i) & \mbox{otherwise}
    \end{array}
\right.$\;
$\delta_{j,i} \longleftarrow y_i - Q_{\theta_j}(x_i, o_i)$\;
}
perform a gradient descent step on $\sum_{i}\delta_{j,i}^2$ with respect to the
network parameters $\theta_j$\;
$j\longleftarrow j+1$\;
$\wD \longleftarrow \wD ~ \backslash ~ \{(x_i,o_i,\rho_i,x'_i,k_i)\}$\;
}
}
\end{algorithm2e*}

\section{SMDP-based Offline RL} \label{sec:smdp-algs}
We introduce offline SMDP methods: a new class of algorithms that operate for variable-timed SMDP settings in an offline manner.  

\subsection{Algorithm Design}

We target offline \textit{value-based} algorithms in this paper. The core technique is as follows: we first update the Bellman target for (value-based) deep RL methods along the same lines as the SMDP Q-learning update rule. Next, we add algorithm-specific tricks to achieve the SMDP version of the algorithm. Specifically, in the MDP scenarios, the $Q$ function is typically approximated using a parameterised function $Q_\theta$, where $\theta$ is a vector of parameters. At each iteration $j$, the parameter vector is trained by minimising a sequence of squared temporal-difference errors:
\begin{align} 
    \nonumber L_j(\theta_j) &\doteq \mathbb{E}_{(x,a,r,x')\sim U(\mathcal{D})} ~ \delta_{j}^2 \\
    \delta_{j} &\doteq r + \gamma \max_{a'} Q_{\theta_j} (x',a') - Q_{\theta_j}(x, a) \label{eq:normal-loss}
\end{align}
where $U(\mathcal{D})$ denotes uniform sampling from the offline data $\mathcal{D}$, and $\delta_{j}$ is the TD error at iteration $j$. 

We first construct a new data set $\mathcal{D}'$ at the option level in the form of $\{(x,o,\rho,x',k)\}$, where $x$ and $x'$ are the states where $o$ is initiated and terminated, respectively; $\rho$ is the \textit{discounted accumulated rewards} in the course of $o$; and $k$ is the length of $o$. We then replace \eqref{eq:normal-loss} with the option error as the following:
\begin{align}
    \nonumber L_j(\theta_j) &\doteq \mathbb{E}_{(x,o,\rho,x',k)\sim U(\mathcal{D}')} ~ \delta_{j}^{2} \\
    \delta_{j} &\doteq \rho + \gamma^k \max_{o'} Q_{\theta_j} (x',o') - Q_{\theta_j}(x, o) \label{eq:option-loss}
\end{align}
\algorithmref{alg:smdp} presents a vanilla SMDP offline RL algorithm based on \eqref{eq:option-loss}. Various more sophisticated algorithms can follow \eqref{eq:option-loss} immediately; we present two basic ones in this section and a more advanced one in the next section. In all such algorithms, the relevant lines of \algorithmref{alg:smdp} should properly be modified. Of note, to mitigate the \textit{value overflow} problem \citep{Fatemi2019-security}, in all of our methods we also clip the $\max_{o'} Q_{\theta_j} (x',o')$ part of \eqref{eq:option-loss} to remain between the minimum and maximum possible returns, when applicable. 

{\flushleft\textbf{SDQN.}} Using the loss function from \eqref{eq:option-loss}, we can then generalize the original Deep Q-networks (DQN) method \citep{mnih2015human}: We introduce \textit{SDQN} by considering a main network $Q_{\theta}$ and a target network $Q_{\theta'}$, which is updated from the main network every $K$ steps. The loss function \eqref{eq:option-loss} is then modified by replacing $\max_{o'} Q_{\theta_j} (x',o')$ with $\max_{o'} Q_{\theta'_j} (x',o')$.

{\flushleft\textbf{SDDQN.}} In a similar way, we can extend the double DQN (DDQN) algorithm \citep{Hasselt2016}, which suppresses over-estimation issues of DQN by decoupling action-selection from maximization in the TD error. Our modification yields the new \textit{SDDQN} variation with the following TD error:
\begin{align}
    \nonumber \delta_j \doteq 
    \rho + \gamma^k Q_{\theta'_j} \left(x',\mbox{argmax}_{o'}Q_{\theta_j}(x',o')\right) - Q_{\theta_j}(x, o)
\end{align}

\subsection{SMDP Batch-Constrained Q-Learning}
One of the challenges of offline RL methods is mitigating the adverse effects of a phenomenon called extrapolation error \citep{fujimoto2019off}. There are a few causes of extrapolation error, such as insufficient and/or imbalanced data, model bias, and training mismatch. In all these cases, off-policy learning methods may yield an arbitrarily inaccurate value estimate $Q(s,a)$ and a sub-optimal policy. To overcome this, Batch-Constrained Q-learning (BCQ) restrains the action space during the training process to actions that are likely to be observed with a probability larger than threshold $\tau$ \citep{fujimoto2019off}. 

Central to BCQ is three core steps: (1) \textit{minimize} the distance of selected actions to the data in the batch, (2) find states where observed data is similar to the batch, and (3) \textit{maximize} the value function based on those. In the continuous setting, BCQ trains a generative model $G_{\omega}(a_i|x)$, commonly a conditional variational auto-encoder \citep{kingma2013-vae, sohn2015-vae}, to estimate the state-conditioned marginal likelihood of observing action $a_i$ at state $s$. Actions are sampled from this network, and the highest-valued action is selected. In the discrete case, which is used in this paper, the same core principles are maintained with a simpler approach \citep{fujimoto2019benchmarking}. $G_{\omega}(a_i|x)$ is instead a behavioural cloning network trained with a cross-entropy loss and used to mask actions whose relative probability are below a threshold $\tau$ (i.e. unlikely to be seen at the given state).

Originally, BCQ maintains four neural networks: two main networks $Q_{\theta_1}$, $Q_{\theta_2}$, and two target networks $Q_{\theta'_1}$, $Q_{\theta'_2}$. To compute the loss for the training of both main networks, the target value for a transition $(x,a,x')$ is given by the following:
\begin{align} \label{eq:bcq:q}
    \nonumber y = r(x,a,x') + \gamma &\max_{a_{i}} [ \lambda\min_{j=1,2} Q_{\theta'_{j}}(x',a_{i}) \\
     &+ (1-\lambda)\max_{j=1,2} Q_{\theta'_{j}}(x',a_{i}) ]
\end{align}
where $a_i$'s are such that $\frac{G_{\omega}(a_{i}|x)}{\max_{\hat{a}} G_{\omega}(\hat{a}'|x)}  > \tau$ for some $\tau$, and $\lambda\in (0,1)$. In the discrete case, BCQ simply uses a double-DQN form as discussed in the previous section. The target networks are updated from the main networks either as a lump update once every $K$ steps, or gradually by $\theta'_{j} = \kappa\theta_{j} + (1-\kappa)\theta'_{j}$ with some fixed $\kappa\in (0,1)$.  

The resulting policy is then as follows: 
\begin{align}
    \nonumber \pi(x) = \mathrm{argmax}_{a'} Q_{\theta} (x, a'), ~
    a' ~ \mbox{s.t.}  ~ \frac{G_{\omega}(a'|x)}{\max_{\hat{a}} G_{\omega}(\hat{a}'|x)}  > \tau
\end{align}

In order to adopt BCQ and construct its SMDP version, which we term \textit{SBCQ}, we need to modify (\ref{eq:bcq:q}) in accordance with (\ref{eq:option-loss}).
Remark that in (\ref{eq:bcq:q}), the entire part in brackets yields $Q$ of the next state, $x'$, on which the max operator applies under the batch-constraint. We therefore need to replace $\gamma$ with $\gamma^k$, $r$ with $\rho$, and replace actions with options. Additionally, the generative model, $G_{\omega}(a_i|x)$, should be replaced with an option-level version, $G_{\omega}(o_i|x)$, and the action generation process should be performed over options instead. The rest of the algorithmic procedures in the original BCQ remains the same.

\subsection{Estimation Requirements}\label{sec:rewards}

In the algorithms presented above, the constructed dataset, $D'$, contains transitions in the form of $(x,o,\rho,x')$. Hence, only the first and last states where $o$ begins and ends are required to be known, not any others during the execution of $o$. However, the rewards at all the \textit{intra} transitions are needed for computing the discounted return $\rho$. 
On the other hand, if the environment is SMDP but modeled as an MDP with aggregation of data at primitive time steps, then we will also require estimating states and actions in addition to rewards at all the primitive steps. This is in general significantly more limiting and is normally error prone, as we further will see in Section \ref{sec:warfarin_mdp}. It is also worth noting that in some problems, domain knowledge may help determine how to compute $\rho$ (even fully accurately), which is rarely the case for states and actions. For example, in goal-oriented tasks, the reward is always zero except when the goal is achieved, where the last option also terminates. Hence, $\rho=0$ for all options except the last one, where $\rho=\gamma^{k-1}r_g$, with $r_g$ being the goal reward.

To estimate the rewards in the warfarin problem in particular, we refer to the medical literature and assume linear changes in subsequent INR measurements, as suggested by \cite{Rosendaal1993AMT}. This is because the goal of warfarin dosing management is to maximize the time we spend in the INR therapeutic range of 2-3, also known as the time in therapeutic range (TTR). To encourage this behaviour, the reward signal is +1 if the subsequent INR is within the therapeutic range and 0 otherwise. To obtain intra transition rewards, we can use linear interpolation to estimate the INR between clinical visits, and use these estimates to determine the reward signals and thus the discounted return $\rho$.

\section{Illustrative Example}
To demonstrate the need for SMDP-based algorithms in time-variable environments, we compare the MDP and SMDP variations for each of the three algorithms presented in Section \ref{sec:smdp-algs} in a simple Minigrid environment \citep{gym_minigrid}. We intentionally design the environment such that only options are accessible to the agent and observations are only provided when options are terminated. Experimental details can be found in the Appendix.

\subsection{Environment Setup}
\begin{figure}[t!]
    \centering
    \includegraphics[width=0.35\linewidth]{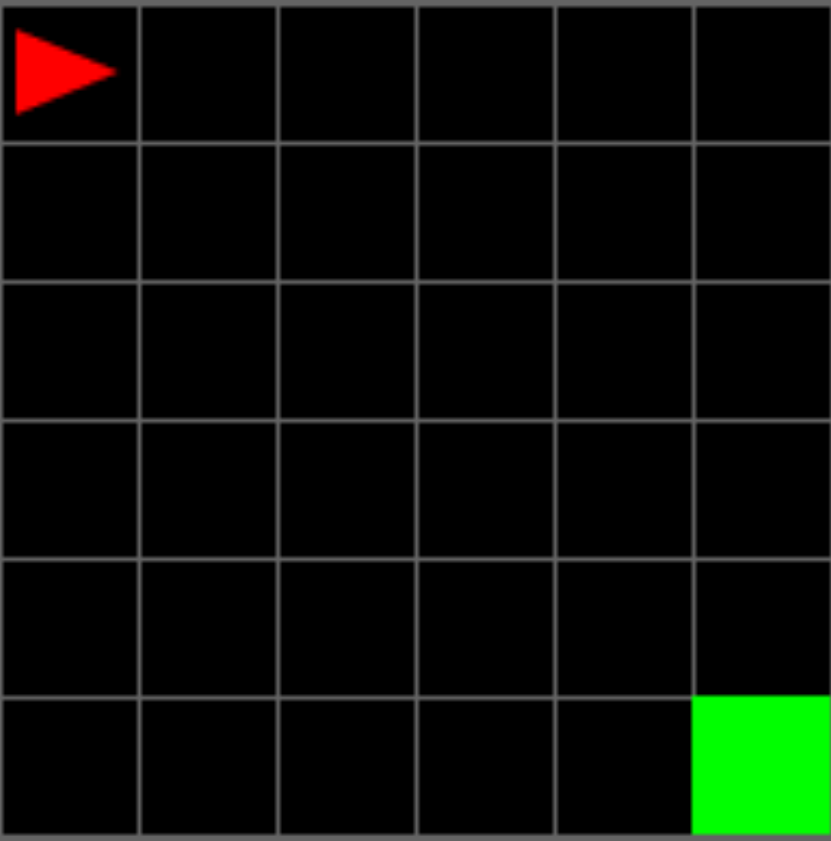}
    \caption{Minigrid 8x8 environment.}
    \label{fig:my_label}
\end{figure}

In the Minigrid, shown in Figure \ref{fig:my_label}, the agent (red triangle) attempts to reach the goal state (green square) to achieve a reward of +10. The \textit{primitive} actions in the environment are: turn left, turn right, move forward, and do nothing. Each option begins by selecting one of the primitive actions, followed by a series of forward steps. The number of steps depends on the agent's location in the grid --- when the agent is in the leftmost column facing downwards, the step size is 4. When the agent is in the top row, the step size is 1. Elsewhere, the step size is 2. To emulate real-world scenarios with intra-transition reward signals, we introduce a -1 reward from crossing the third row, unless we are in the rightmost column, where no negative reward exists. 

\subsection{MDP versus SMDP}\label{sec:smdp_vs_mdp_online}
\begin{figure*}
    \centering
    \includegraphics{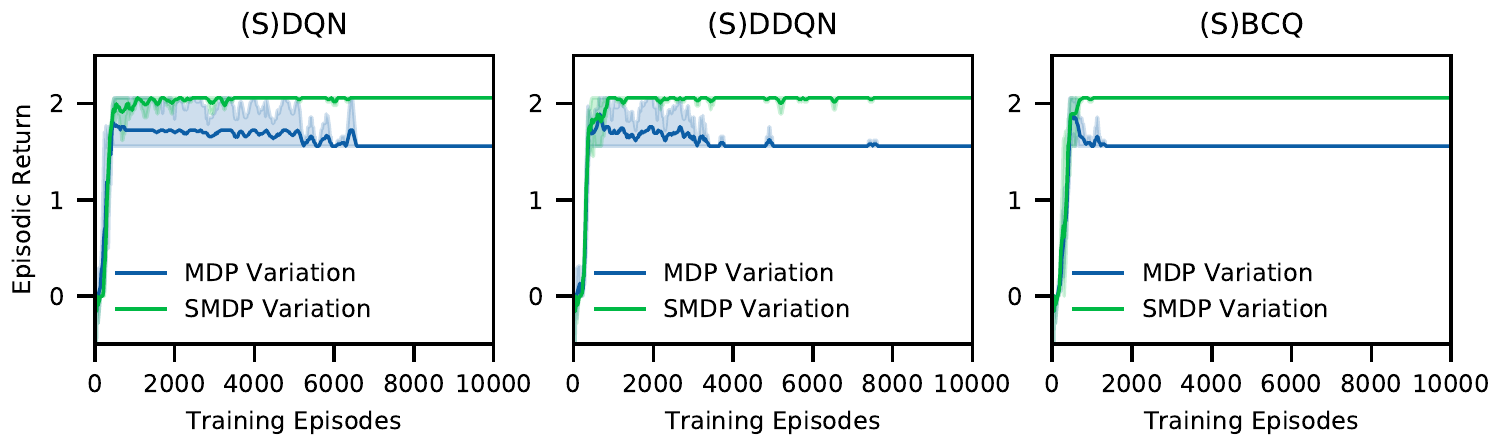}
    \caption{Episodic returns across three random seeds for (S)DQN, and (S)DDQN, and (S)BCQ.}
    \label{fig:learning_curves}
\end{figure*}

The first experiment compares the MDP and SMDP variations in an online setting, where the agents are trained with the same exploration scheme. The learning curves (Figure \ref{fig:learning_curves}) demonstrate that for all three algorithms, the traditional MDP variations yield a sub-optimal policy, whereas the SMDP variations all converge to the optimal policy.  
This is because the traditional MDP algorithms consider option-level transitions as primitive transitions. They do not account for the steps taken inside the options, thus improperly discounting the goal reward. In other words, they choose to reach the goal (mistakenly) faster at the expense of collecting the negative reward from the third row. In contrast, SMDP algorithms correctly compute the return, and as a result, choose the optimal trajectory that achieves a higher return. It is worth noting that the MDP algorithms were able to converge due to the simplicity of the example, but in general, MDP algorithms in SMDP environments may not converge at all (even to incorrect values) due to nonstationary behaviours.

\subsection{Offline Learning}
The next experiment compares the three SMDP variations in an offline setting. Two factors in offline RL that can pose challenges are the quality and size of the dataset. To analyze how these affect SDQN, SDDQN, and SBCQ, we create datasets consisting each of 100, 1,000, and 10,000 transitions. 
Trajectories in each dataset include certain percentages of optimal and sub-optimal options. We introduce two sub-optimal options that can be chosen at any step: the second-best option (25\% of the time) and a random option (10\%, 25\%, and 50\% of the time). This setup is intended to loosely resemble real-world scenarios, such as healthcare treatment management problems, where treatment decisions are not always ``optimal'' but are not fully random either. 
We train the agent using each of these datasets, and for each configuration, the network which minimizes the validation loss is used to evaluate on the test set. The test episodes are fixed across all experiments, and consists of 10 randomly selected initial starting points to test the agent's learned policy more thoroughly.

\begin{figure*}[h!]
    \centering
    \includegraphics{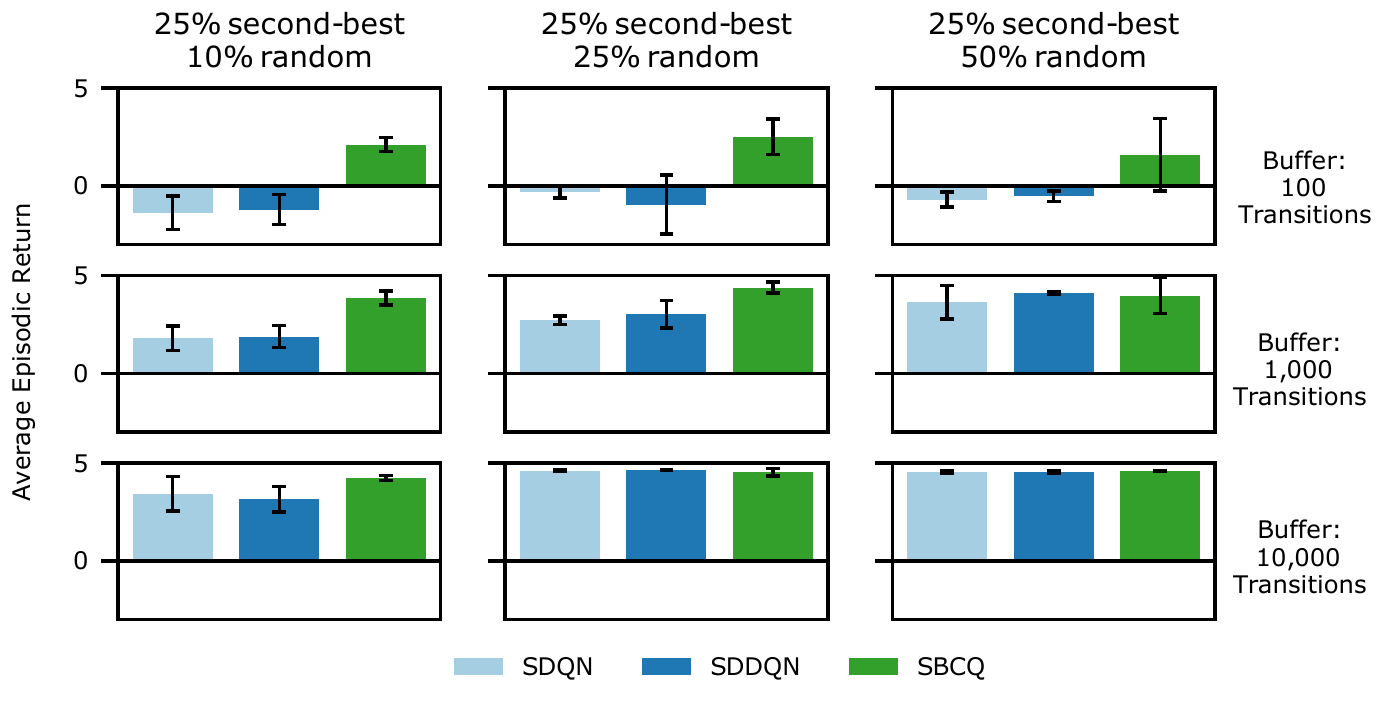}
    \caption{Episodic returns averaged across test cases for each dataset configuration and each of SDQN, SDDQN, and SBCQ. The environment here is slightly different from Figure \ref{fig:learning_curves} and explained in the Appendix.}
    \label{fig:buffer_analysis}
\end{figure*}

Figure \ref{fig:buffer_analysis} shows the episodic returns, averaged across the 10 test episodes for each algorithm and dataset combination. We highlight two important conclusions: first, at small dataset sizes, SBCQ greatly outperforms SDQN and SDDQN, particularly when the dataset has more optimal decisions. Second, when the buffer size increases, SDQN and SDDQN both improve performance but are at best equivalent to SBCQ. SBCQ appears to learn more robustly with a small and noisy dataset than SDQN or SDDQN. As a result, SBCQ is expected to induce a better estimate than SDQN or SDDQN for many applications of interest, such as healthcare, where datasets may be limited in size and quality.

\section{Warfarin Dosing}
In this section, we consider the warfarin dosing optimization setting. As described in the introduction, warfarin is an anti-coagulant commonly prescribed to patients with atrial fibrillation to reduce their risk of stroke, with dose management targeting an INR level between 2 and 3 (known as the therapeutic range).  

\subsection{Data}
The warfarin dataset consists of data for 29,272 adult patients across four randomized control trials: ARISTOTLE \citep{aristotle}, ENGAGE AF \citep{engage}, ROCKET AF \citep{rocket}, and RELY \citep{rely}. The dataset contains records for each clinical visit, which includes the measured INR and the prescribed warfarin dose, as well as records of adverse clinical events. This is complemented by patient medical history and demographic information. We filter for patients who did not have warfarin dose or INR data, as well as for patients who had weekly doses exceeding 140mg. The final cohort, after preprocessing, consists of 28,444 patients, with trajectories that are on average $26.9 \pm 13.1$ steps and $604.5 \pm 318.1$ days.

The state space is comprised of the measured INR values, the prescribed warfarin dose, and the patient health indicators. The indicators are suggested by clinicians based on clinical expertise, and are sex, patient geography, medications at randomization (amiodarone, aspirin), and medical history (diabetes, myocardial infarction, hypertension, and smoking). There are seven options available to clinicians: decreasing or increasing the dose by increments of $>$20\%, 10-20\%, or 0-10\%, or maintaining the current dose. As discussed in Section \ref{sec:rewards}, we use linear interpolation to estimate the INR for each intra transition and assign a reward of +1 if the INR is within the therapeutic range of 2-3 and 0 otherwise. The $\rho$ is the discounted return of these rewards over the intermediate time steps prior to the next observation. 

To split the data, we randomly sample half of the ARISTOTLE data to create the test dataset. The validation dataset then randomly samples from the remaining ARISTOTLE and RELY data, and the rest of the data is used in training. An important metric in warfarin dosing is the time in therapeutic range (TTR). We estimate the TTR by aggregating the TTRs of the trajectories which have an agreement between the policy and clinician actions that exceed $85\%$ of the training trajectories' agreements. The model with the highest estimated TTR on the validation dataset is chosen for evaluation on the test dataset. Technical details of data preparation and splitting are presented in the Appendix.

\subsection{MDP versus SMDP}\label{sec:warfarin_mdp}

As discussed in the introduction, timing is not consistent in many practical situations, and modelling these as MDPs, with or without artificial reshaping of the data, may lead to unreliable results. The warfarin problem is a salient example. The duration between subsequent visits range from one day to three months (mean 23 $\pm$ 15 days). Importantly, the time between earlier visits may be short (1-2 days), with many transitions and option decisions during this phase. Data aggregation of weekly data points, for example, would lose the information from these entries. Hence, data aggregation must be done at a more fine-grain level, which results in unclear dynamics and numerous fake points, especially in longer visit durations (even for a single patient). Another spurious solution is to introduce time into the state space. In principle, this should make the state informative and seemingly solve the issue. However, with limited data, adding time to the state simply makes the observation space far too sparse and not learnable. We found that in practice, adding time significantly increases the training loss with almost no convergence, as expected.

\begin{figure}[t!]
    \centering
    \includegraphics{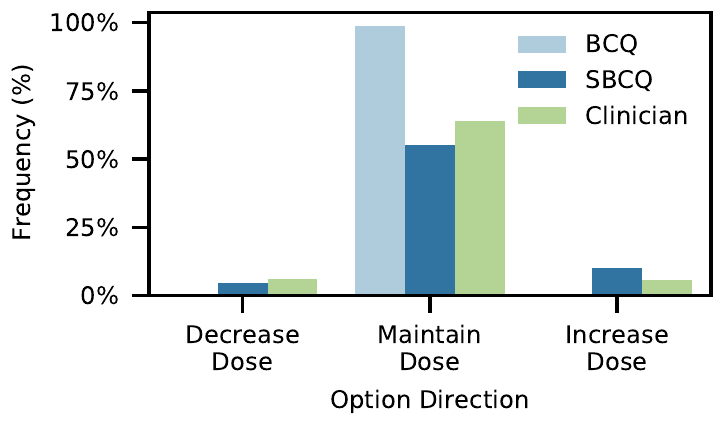}
    \caption{Options chosen by the interpolated MDP formulation of BCQ, SBCQ, and the clinicians for all test states, aggregated by option direction.}
    \label{fig:action_hist}
\end{figure}

Here, we examine the effect of artificial data aggregation in the warfarin setting using time steps of one day. We use linear interpolation to define the state where data is not available. This is similar to the process for computing the discounted return of an option. However, a critical distinction is that the interpolated INR is used here in both the state and reward, as opposed to the return estimation process, which only uses the results to calculate the rewards. 
Figure \ref{fig:action_hist} illustrates how BCQ using this MDP formulation yields a policy that frequently chooses to maintain the current dose. This is due to the imbalanced option space with excessive repetition of the option ``Maintain Dose'' after data aggregation. The SBCQ algorithm, on the other hand, is able to learn a policy that is more nuanced and noticeably follows similar option directions as the clinician with slight preference towards increasing the dosage.

\begin{figure}[t!]
    \centering
    \includegraphics{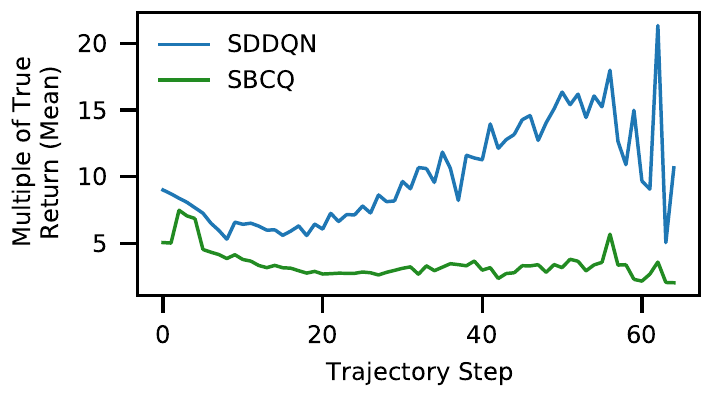}
    \caption{Deviation of SDDQN and SBCQ from the true returns (as a multiple of the true returns), averaged over test patients at each step along their trajectories.}
    \label{fig:mean_qvalues}
\end{figure}

\subsection{SDDQN versus SBCQ}
We demonstrate that SBCQ is superior to SDDQN in the warfarin setting. In particular, we investigate the problem of over-estimation. We take the difference between the learned Q-values for both SDDQN and SBCQ against the true returns for each point along the patient's trajectory. We then normalize the difference by the true return and show the mean across all test patients for each trajectory step, illustrated in Figure \ref{fig:mean_qvalues}. We limit the trajectory length to 65 steps, as more than 99.5\% of the test patient trajectories satisfy this condition.
Figure \ref{fig:mean_qvalues} demonstrates that A) SBCQ incurs nearly an order of magnitude lower over-estimation, and B) the over-estimation is largely consistent along steps, hence it may be less destructive in terms of recovering the optimal policy. Remark that in an ideal case, where over-estimation is a fixed-value shift, the optimal policy becomes fully recoverable.

\section{Concluding Remarks}
In this paper, we focused on offline RL settings with inconsistent timing, which conflate two orthogonal characteristics: being non-Markovian and being offline. We appealed to the Semi-Markov Decision Process (SMDP) theory and presented a formal methodology to modify any given value-based offline RL method to its SMDP variation. This way, we formally addressed the time-varying issue, while keeping the algorithmic machinery consistent with the contemporary offline RL literature. Consequently, with the advent of more preferable offline RL methods, we can readily translate them to their SMDP counterparts. 

We further leveraged the design space that our methodology provides, and proposed three algorithms: SDQN, SDDQN, and SBCQ. To examine the resultant algorithms, we experimentally demonstrated that these algorithms converged to the optimal policy in a time-varying online environment, whereas their MDP counterparts converged to sub-optimal policies. We then switched to offline settings and demonstrated that SBCQ was considerably more robust than SDQN or SDDQN, particularly when trained using smaller and/or noisier datasets, as is the case in many practical applications, such as healthcare. Finally, we presented a real-world scenario pertaining to warfarin dosing management. This problem remarkably manifests both the time-varying and offline properties. We demonstrated that the MDP framing of the problem was unable to learn a meaningful policy. Turning to the SMDP algorithms, we then showed that SBCQ was less susceptible to over-estimation than SDDQN with a significant margin. 

To conclude, both the formal results and our experimental investigations suggest that SMDP modelling is required in certain practical applications such as healthcare, resulting in robust and more reliable policies, especially when dealing with limited offline data.

\section*{Institutional Review Board (IRB)}
Approval from the Duke University Institutional Review Board as well as institutional review boards at collaborating centers (where required) was obtained prior to receipt of any outside data. Inclusion of this element of PHI in the combined COMBINE AF is registered with PROSPERO (CRD42020178771).

\acks{We are indebted to our colleagues, who provided thoughtful insights and feedback, with special thanks to Taylor W. Killian for his advice during the project scoping process. We would also like to express our gratitude to the COMBINE AF investigators for providing the data and assisting us through any data questions. We also kindly thank the anonymous reviewers for their suggested improvements. 

This research was supported in part by Microsoft Research, a CIFAR Azrieli Global Scholar Chair, a
Canada Research Council Chair, and an NSERC Discovery Grant.
}

\bibliography{jmlr-sample}

\begin{thebibliography}{63}
\providecommand{\natexlab}[1]{#1}
\providecommand{\url}[1]{\texttt{#1}}
\expandafter\ifx\csname urlstyle\endcsname\relax
  \providecommand{\doi}[1]{doi: #1}\else
  \providecommand{\doi}{doi: \begingroup \urlstyle{rm}\Url}\fi

\bibitem[Aalen et~al.(2008)Aalen, Borgan, and Gjessing]{Aalen2008}
Odd~O. Aalen, {\O}rnulf Borgan, and H{\aa}kon~K. Gjessing.
\newblock \emph{Survival and Event History Analysis}.
\newblock Springer New York, 2008.
\newblock \doi{10.1007/978-0-387-68560-1}.
\newblock URL \url{https://doi.org/10.1007/978-0-387-68560-1}.

\bibitem[Agarwal et~al.(2020)Agarwal, Schuurmans, and
  Norouzi]{agarwal2020optimistic}
Rishabh Agarwal, Dale Schuurmans, and Mohammad Norouzi.
\newblock An optimistic perspective on offline reinforcement learning.
\newblock In \emph{International Conference on Machine Learning}, pages
  104--114. PMLR, 2020.

\bibitem[Bellinger et~al.(2020)Bellinger, Coles, Crowley, and
  Tamblyn]{Bellinger2020}
Colin Bellinger, Rory Coles, Mark Crowley, and Isaac Tamblyn.
\newblock Reinforcement learning in a physics-inspired semi-markov environment.
\newblock April 2020.

\bibitem[Bertsekas and Tsitsiklis(1996)]{bertsekas_neuro}
Dimitri~P. Bertsekas and John~N. Tsitsiklis.
\newblock \emph{Neuro-Dynamic Programming}.
\newblock Athena Scientific, 1st edition, 1996.
\newblock ISBN 1886529108.

\bibitem[Bradtke and Duff(1994)]{Bradtke_undated-tf}
Steven~J. Bradtke and Michael~O. Duff.
\newblock Reinforcement learning methods for continuous-time markov decision
  problems.
\newblock In \emph{Proceedings of the 7th International Conference on Neural
  Information Processing Systems}, NIPS'94, page 393–400, Cambridge, MA, USA,
  1994. MIT Press.

\bibitem[Carnicelli et~al.(2021)Carnicelli, Hong, Giugliano, Connolly,
  Eikelboom, Patel, Wallentin, Morrow, Wojdyla, Hua, Hohnloser, Oldgren, Ruff,
  Piccini, Lopes, Alexander, and Granger]{Carnicelli2021}
Anthony~P Carnicelli, Hwanhee Hong, Robert~P Giugliano, Stuart~J Connolly, John
  Eikelboom, Manesh~R Patel, Lars Wallentin, David~A Morrow, Daniel Wojdyla,
  Kaiyuan Hua, Stefan~H Hohnloser, Jonas Oldgren, Christian~T Ruff, Jonathan~P
  Piccini, Renato~D Lopes, John~H Alexander, and Christopher~B Granger.
\newblock Individual patient data from the pivotal randomized controlled trials
  of non-vitamin k antagonist oral anticoagulants in patients with atrial
  fibrillation ({COMBINE} {AF}): Design and rationale.
\newblock \emph{American Heart Journal}, 233:\penalty0 48--58, March 2021.
\newblock \doi{10.1016/j.ahj.2020.12.002}.
\newblock URL \url{https://doi.org/10.1016/j.ahj.2020.12.002}.

\bibitem[Chevalier-Boisvert et~al.(2018)Chevalier-Boisvert, Willems, and
  Pal]{gym_minigrid}
Maxime Chevalier-Boisvert, Lucas Willems, and Suman Pal.
\newblock Minimalistic gridworld environment for openai gym.
\newblock \url{https://github.com/maximecb/gym-minigrid}, 2018.

\bibitem[Chrisman(1994)]{Chrisman1994ReasoningAP}
Lonnie Chrisman.
\newblock Reasoning about probabilistic actions at multiple levels of
  granularity.
\newblock 1994.

\bibitem[Connolly et~al.(2009)Connolly, Ezekowitz, Yusuf, Eikelboom, Oldgren,
  Parekh, Pogue, Reilly, Themeles, Varrone, Wang, Alings, Xavier, Zhu, Diaz,
  Lewis, Darius, Diener, Joyner, and Wallentin]{rely}
Stuart~J. Connolly, Michael~D. Ezekowitz, Salim Yusuf, John Eikelboom, Jonas
  Oldgren, Amit Parekh, Janice Pogue, Paul~A. Reilly, Ellison Themeles, Jeanne
  Varrone, Susan Wang, Marco Alings, Denis Xavier, Jun Zhu, Rafael Diaz,
  Basil~S. Lewis, Harald Darius, Hans-Christoph Diener, Campbell~D. Joyner, and
  Lars Wallentin.
\newblock Dabigatran versus warfarin in patients with atrial fibrillation.
\newblock 361\penalty0 (12):\penalty0 1139--1151, September 2009.
\newblock \doi{10.1056/nejmoa0905561}.
\newblock URL \url{https://doi.org/10.1056/nejmoa0905561}.

\bibitem[Fatemi et~al.(2019)Fatemi, Sharma, Van~Seijen, and
  Kahou]{Fatemi2019-security}
Mehdi Fatemi, Shikhar Sharma, Harm Van~Seijen, and Samira~Ebrahimi Kahou.
\newblock Dead-ends and secure exploration in reinforcement learning.
\newblock In Kamalika Chaudhuri and Ruslan Salakhutdinov, editors,
  \emph{Proceedings of the 36th International Conference on Machine Learning},
  volume~97 of \emph{Proceedings of Machine Learning Research}, pages
  1873--1881, Long Beach, California, USA, 2019. PMLR.

\bibitem[Fatemi et~al.(2021)Fatemi, Killian, Subramanian, and
  Ghassemi]{fatemi2021medical}
Mehdi Fatemi, Taylor~W. Killian, Jayakumar Subramanian, and Marzyeh Ghassemi.
\newblock Medical dead-ends and learning to identify high-risk states and
  treatments, 2021.

\bibitem[Fran{\c{c}}ois-Lavet et~al.(2019)Fran{\c{c}}ois-Lavet, Rabusseau,
  Pineau, Ernst, and Fonteneau]{francois2019overfitting}
Vincent Fran{\c{c}}ois-Lavet, Guillaume Rabusseau, Joelle Pineau, Damien Ernst,
  and Raphael Fonteneau.
\newblock On overfitting and asymptotic bias in batch reinforcement learning
  with partial observability.
\newblock \emph{Journal of Artificial Intelligence Research}, 65:\penalty0
  1--30, 2019.

\bibitem[Fujimoto et~al.(2019{\natexlab{a}})Fujimoto, Conti, Ghavamzadeh, and
  Pineau]{fujimoto2019benchmarking}
Scott Fujimoto, Edoardo Conti, Mohammad Ghavamzadeh, and Joelle Pineau.
\newblock Benchmarking batch deep reinforcement learning algorithms,
  2019{\natexlab{a}}.

\bibitem[Fujimoto et~al.(2019{\natexlab{b}})Fujimoto, Meger, and
  Precup]{fujimoto2019off}
Scott Fujimoto, David Meger, and Doina Precup.
\newblock Off-policy deep reinforcement learning without exploration.
\newblock In Kamalika Chaudhuri and Ruslan Salakhutdinov, editors,
  \emph{Proceedings of the 36th International Conference on Machine Learning},
  volume~97 of \emph{Proceedings of Machine Learning Research}, pages
  2052--2062. PMLR, 09--15 Jun 2019{\natexlab{b}}.

\bibitem[Futoma et~al.(2017)Futoma, Hariharan, Heller, Sendak, Brajer, Clement,
  Bedoya, and O’Brien]{futoma2017improved}
Joseph Futoma, Sanjay Hariharan, Katherine Heller, Mark Sendak, Nathan Brajer,
  Meredith Clement, Armando Bedoya, and Cara O’Brien.
\newblock An improved multi-output gaussian process rnn with real-time
  validation for early sepsis detection.
\newblock In \emph{Machine Learning for Healthcare Conference}, pages 243--254,
  2017.

\bibitem[Ghassemi et~al.(2014)Ghassemi, Naumann, Doshi-Velez, Brimmer, Joshi,
  Rumshisky, and Szolovits]{ghassemi2014unfolding}
Marzyeh Ghassemi, Tristan Naumann, Finale Doshi-Velez, Nicole Brimmer, Rohit
  Joshi, Anna Rumshisky, and Peter Szolovits.
\newblock Unfolding physiological state: Mortality modelling in intensive care
  units.
\newblock In \emph{Proceedings of the 20th ACM SIGKDD international conference
  on Knowledge discovery and data mining}, pages 75--84. ACM, 2014.

\bibitem[Ghassemi et~al.(2015)Ghassemi, Pimentel, Naumann, Brennan, Clifton,
  Szolovits, and Feng]{ghassemi2015multivariate}
Marzyeh Ghassemi, Marco~AF Pimentel, Tristan Naumann, Thomas Brennan, David~A
  Clifton, Peter Szolovits, and Mengling Feng.
\newblock A multivariate timeseries modeling approach to severity of illness
  assessment and forecasting in icu with sparse, heterogeneous clinical data.
\newblock In \emph{Proc. Twenty-Ninth AAAI Conf. on Artificial Intelligence},
  2015.

\bibitem[Ghassemi et~al.(2017)Ghassemi, Wu, Hughes, and
  Doshi-Velez]{ghassemi2017predicting}
Marzyeh Ghassemi, Mike Wu, Michael Hughes, and Finale Doshi-Velez.
\newblock Predicting intervention onset in the icu with switching state space
  models.
\newblock In \emph{Proceedings of the AMIA Summit on Clinical Research
  Informatics (CRI)}, volume 2017. American Medical Informatics Association,
  2017.

\bibitem[Giugliano et~al.(2013)Giugliano, Ruff, Braunwald, Murphy, Wiviott,
  Halperin, Waldo, Ezekowitz, Weitz, {\v{S}}pinar, Ruzyllo, Ruda, Koretsune,
  Betcher, Shi, Grip, Patel, Patel, Hanyok, Mercuri, and Antman]{engage}
Robert~P. Giugliano, Christian~T. Ruff, Eugene Braunwald, Sabina~A. Murphy,
  Stephen~D. Wiviott, Jonathan~L. Halperin, Albert~L. Waldo, Michael~D.
  Ezekowitz, Jeffrey~I. Weitz, Jind{\v{r}}ich {\v{S}}pinar, Witold Ruzyllo,
  Mikhail Ruda, Yukihiro Koretsune, Joshua Betcher, Minggao Shi, Laura~T. Grip,
  Shirali~P. Patel, Indravadan Patel, James~J. Hanyok, Michele Mercuri, and
  Elliott~M. Antman.
\newblock Edoxaban versus warfarin in patients with atrial fibrillation.
\newblock 369\penalty0 (22):\penalty0 2093--2104, November 2013.
\newblock \doi{10.1056/nejmoa1310907}.
\newblock URL \url{https://doi.org/10.1056/nejmoa1310907}.

\bibitem[Granger et~al.(2011)Granger, Alexander, McMurray, Lopes, Hylek, Hanna,
  Al-Khalidi, Ansell, Atar, Avezum, Bahit, Diaz, Easton, Ezekowitz, Flaker,
  Garcia, Geraldes, Gersh, Golitsyn, Goto, Hermosillo, Hohnloser, Horowitz,
  Mohan, Jansky, Lewis, Lopez-Sendon, Pais, Parkhomenko, Verheugt, Zhu, and
  Wallentin]{aristotle}
Christopher~B. Granger, John~H. Alexander, John~J.V. McMurray, Renato~D. Lopes,
  Elaine~M. Hylek, Michael Hanna, Hussein~R. Al-Khalidi, Jack Ansell, Dan Atar,
  Alvaro Avezum, M.~Cecilia Bahit, Rafael Diaz, J.~Donald Easton, Justin~A.
  Ezekowitz, Greg Flaker, David Garcia, Margarida Geraldes, Bernard~J. Gersh,
  Sergey Golitsyn, Shinya Goto, Antonio~G. Hermosillo, Stefan~H. Hohnloser,
  John Horowitz, Puneet Mohan, Petr Jansky, Basil~S. Lewis, Jose~Luis
  Lopez-Sendon, Prem Pais, Alexander Parkhomenko, Freek~W.A. Verheugt, Jun Zhu,
  and Lars Wallentin.
\newblock Apixaban versus warfarin in patients with atrial fibrillation.
\newblock \emph{New England Journal of Medicine}, 365\penalty0 (11):\penalty0
  981--992, 2011.
\newblock \doi{10.1056/NEJMoa1107039}.
\newblock URL \url{https://doi.org/10.1056/NEJMoa1107039}.
\newblock PMID: 21870978.

\bibitem[Hasselt et~al.(2016)Hasselt, Guez, and Silver]{Hasselt2016}
Hado~van Hasselt, Arthur Guez, and David Silver.
\newblock Deep reinforcement learning with double q-learning.
\newblock In \emph{Proceedings of the Thirtieth AAAI Conference on Artificial
  Intelligence}, AAAI’16, page 2094–2100. AAAI Press, 2016.

\bibitem[Henry et~al.(2015)Henry, Hager, Pronovost, and
  Saria]{henry2015targeted}
Katharine~E Henry, David~N Hager, Peter~J Pronovost, and Suchi Saria.
\newblock A targeted real-time early warning score (trewscore) for septic
  shock.
\newblock \emph{Science translational medicine}, 7\penalty0 (299):\penalty0
  299ra122--299ra122, 2015.

\bibitem[Jaques et~al.(2019)Jaques, Ghandeharioun, Shen, Ferguson, Lapedriza,
  Jones, Gu, and Picard]{jaques2019way}
Natasha Jaques, Asma Ghandeharioun, Judy~Hanwen Shen, Craig Ferguson, Agata
  Lapedriza, Noah Jones, Shixiang Gu, and Rosalind Picard.
\newblock Way off-policy batch deep reinforcement learning of implicit human
  preferences in dialog.
\newblock \emph{arXiv preprint arXiv:1907.00456}, 2019.

\bibitem[Kingma and Welling(2013)]{kingma2013-vae}
Diederik~P Kingma and Max Welling.
\newblock Auto-encoding variational bayes, 2013.

\bibitem[Komorowski et~al.(2018)Komorowski, Celi, Badawi, Gordon, and
  Faisal]{komorowski2018artificial}
Matthieu Komorowski, Leo~A Celi, Omar Badawi, Anthony~C Gordon, and A~Aldo
  Faisal.
\newblock The artificial intelligence clinician learns optimal treatment
  strategies for sepsis in intensive care.
\newblock \emph{Nature medicine}, 24\penalty0 (11):\penalty0 1716--1720, 2018.

\bibitem[Kushner and Yin(2003)]{kushner2003}
Harold Kushner and George Yin.
\newblock \emph{Stochastic Approximation and Recursive Algorithms and
  Applications}.
\newblock Springer-Verlag, 2003.
\newblock \doi{10.1007/b97441}.

\bibitem[Lange et~al.(2012)Lange, Gabel, and Riedmiller]{lange2012batch}
Sascha Lange, Thomas Gabel, and Martin Riedmiller.
\newblock Batch reinforcement learning.
\newblock In \emph{Reinforcement learning}, pages 45--73. Springer, 2012.

\bibitem[Levine et~al.(2020)Levine, Kumar, Tucker, and Fu]{Levine2020}
Sergey Levine, Aviral Kumar, George Tucker, and Justin Fu.
\newblock Offline reinforcement learning: Tutorial, review, and perspectives on
  open problems.
\newblock May 2020.

\bibitem[Li et~al.(2019)Li, Komorowski, and Faisal]{li2019optimizing}
Luchen Li, Matthieu Komorowski, and Aldo~A Faisal.
\newblock Optimizing sequential medical treatments with auto-encoding heuristic
  search in {POMDP}s.
\newblock \emph{arXiv preprint arXiv:1905.07465}, 2019.

\bibitem[Mnih et~al.(2015)Mnih, Kavukcuoglu, Silver, Rusu, Veness, Bellemare,
  Graves, Riedmiller, Fidjeland, Ostrovski, Petersen, Beattie, Sadik,
  Antonoglou, King, Kumaran, Wierstra, Legg, and Hassabis]{mnih2015human}
Volodymyr Mnih, Koray Kavukcuoglu, David Silver, Andrei~A. Rusu, Joel Veness,
  Marc~G. Bellemare, Alex Graves, Martin Riedmiller, Andreas~K. Fidjeland,
  Georg Ostrovski, Stig Petersen, Charles Beattie, Amir Sadik, Ioannis
  Antonoglou, Helen King, Dharshan Kumaran, Daan Wierstra, Shane Legg, and
  Demis Hassabis.
\newblock Human-level control through deep reinforcement learning.
\newblock \emph{Nature}, 518\penalty0 (7540):\penalty0 529--533, 2015.

\bibitem[Nieuwlaat et~al.(2012)Nieuwlaat, Hubers, Spyropoulos, Eikelboom,
  Connolly, Spall, Schulze, Cuddy, Stehouwer, Schulman, and
  Connolly]{Nieuwlaat2012RandomisedCO}
Robby Nieuwlaat, Lowiek~M. Hubers, Alex~C. Spyropoulos, John~W Eikelboom,
  Benjamin~J Connolly, Harriette G. C.~Van Spall, Karleen~M Schulze, Spencer~M
  Cuddy, Alexander~C. Stehouwer, Sam Schulman, and Stuart~J. Connolly.
\newblock Randomised comparison of a simple warfarin dosing algorithm versus a
  computerised anticoagulation management system for control of warfarin
  maintenance therapy.
\newblock \emph{Thrombosis and haemostasis}, 108 6:\penalty0 1228--35, 2012.

\bibitem[Parr(1998)]{Parr1998-au}
Ronald~Edward Parr.
\newblock \emph{Hierarchical Control and Learning for Markov Decision
  Processes}.
\newblock PhD thesis, University of California, Berkeley, 1998.

\bibitem[Patel et~al.(2011)Patel, Mahaffey, Garg, Pan, Singer, Hacke,
  Breithardt, Halperin, Hankey, Piccini, Becker, Nessel, Paolini, Berkowitz,
  Fox, and and]{rocket}
Manesh~R. Patel, Kenneth~W. Mahaffey, Jyotsna Garg, Guohua Pan, Daniel~E.
  Singer, Werner Hacke, G\"{u}nter Breithardt, Jonathan~L. Halperin, Graeme~J.
  Hankey, Jonathan~P. Piccini, Richard~C. Becker, Christopher~C. Nessel,
  John~F. Paolini, Scott~D. Berkowitz, Keith~A.A. Fox, and Robert M.~Califf
  and.
\newblock Rivaroxaban versus warfarin in nonvalvular atrial fibrillation.
\newblock 365\penalty0 (10):\penalty0 883--891, September 2011.
\newblock \doi{10.1056/nejmoa1009638}.
\newblock URL \url{https://doi.org/10.1056/nejmoa1009638}.

\bibitem[Peng et~al.(2018)Peng, Ding, Wihl, Gottesman, Komorowski, Lehman,
  Ross, Faisal, and Doshi-Velez]{peng2018improving}
Xuefeng Peng, Yi~Ding, David Wihl, Omer Gottesman, Matthieu Komorowski,
  Li-wei~H Lehman, Andrew Ross, Aldo Faisal, and Finale Doshi-Velez.
\newblock Improving sepsis treatment strategies by combining deep and
  kernel-based reinforcement learning.
\newblock In \emph{AMIA Annual Symposium Proceedings}, volume 2018, page 887.
  American Medical Informatics Association, 2018.

\bibitem[Pirmohamed(2018)]{warfarin_intro}
Munir Pirmohamed.
\newblock Warfarin: The end of the end of one size fits all therapy.
\newblock In \emph{Journal of Personalized Medicine}, volume~8, 28 June 2018.

\bibitem[Precup and Sutton(1997)]{Precup97multitimemodels}
Doina Precup and Richard~S. Sutton.
\newblock Multi-time models for reinforcement learning.
\newblock In \emph{Proceedings of the ICML'97 Workshop on Modeling in
  Reinforcement Learning}, 1997.

\bibitem[Precup et~al.(1997)Precup, Sutton, and Singh]{Precup97macro}
Doina Precup, Richard~S. Sutton, and Satinder Singh.
\newblock Planning with closed-loop macro actions.
\newblock In \emph{In Working notes of the 1997 AAAI Fall Symposium on
  Model-directed Autonomous Systems}, pages 70--76. MIT Press, 1997.

\bibitem[Precup et~al.(1998)Precup, Sutton, and Singh]{Pericup1998options}
Doina Precup, Richard~S. Sutton, and Satinder Singh.
\newblock Theoretical results on reinforcement learning with temporally
  abstract options.
\newblock In Claire N{\'e}dellec and C{\'e}line Rouveirol, editors,
  \emph{Machine Learning: ECML-98}, pages 382--393, Berlin, Heidelberg, 1998.
  Springer Berlin Heidelberg.

\bibitem[Puterman(1994)]{puterman_book}
Martin Puterman.
\newblock \emph{Markov Decision Processes: Discrete Stochastic Dynamic
  Programming}.
\newblock John Wiley \& Sons, Inc., 1994.
\newblock ISBN 9780471619772.

\bibitem[Raghu et~al.(2017)Raghu, Komorowski, Celi, Szolovits, and
  Ghassemi]{raghu2017continuous}
Aniruddh Raghu, Matthieu Komorowski, Leo~Anthony Celi, Peter Szolovits, and
  Marzyeh Ghassemi.
\newblock Continuous state-space models for optimal sepsis treatment-a deep
  reinforcement learning approach.
\newblock \emph{JMLR (Journal of Machine Learning Research): MLHC Conference
  Proceedings}, 2017.

\bibitem[Rebba et~al.(2006)Rebba, Mahadevan, and Huang]{rebba2006validation}
Ramesh Rebba, Sankaran Mahadevan, and Shuping Huang.
\newblock Validation and error estimation of computational models.
\newblock \emph{Reliability Engineering \&amp; System Safety}, 91\penalty0
  (10-11):\penalty0 1390--1397, 2006.

\bibitem[Ring(1991)]{RING1991343}
Mark Ring.
\newblock Incremental development of complex behaviors through automatic
  construction of sensory-motor hierarchies.
\newblock In Lawrence~A. Birnbaum and Gregg~C. Collins, editors, \emph{Machine
  Learning Proceedings 1991}, pages 343--347. Morgan Kaufmann, San Francisco
  (CA), 1991.
\newblock ISBN 978-1-55860-200-7.
\newblock \doi{https://doi.org/10.1016/B978-1-55860-200-7.50071-4}.

\bibitem[Rosendaal et~al.(1993)Rosendaal, Cannegieter, van~der Meer, and
  Bri{\"e}t]{Rosendaal1993AMT}
Frits~Richard Rosendaal, Suzanne~C. Cannegieter, Felix J.~M. van~der Meer, and
  Ernest Bri{\"e}t.
\newblock A method to determine the optimal intensity of oral anticoagulant
  therapy.
\newblock \emph{Thrombosis and haemostasis}, 69 3:\penalty0 236--9, 1993.

\bibitem[Saria(2018)]{saria2018individualized}
Suchi Saria.
\newblock Individualized sepsis treatment using reinforcement learning.
\newblock \emph{Nature medicine}, 24\penalty0 (11):\penalty0 1641--1642, 2018.

\bibitem[Schmidhuber(1991)]{Schmidhuber91neuralsequence}
J{\"u}rgen Schmidhuber.
\newblock Neural sequence chunkers.
\newblock Technical report, Technische Universitat Munchen, 1991.

\bibitem[Schmoll and Schubert(2020)]{Schmoll2020}
Sebastian Schmoll and Matthias Schubert.
\newblock Semi-markov reinforcement learning for stochastic resource
  collection.
\newblock In Christian Bessiere, editor, \emph{Proceedings of the Twenty-Ninth
  International Joint Conference on Artificial Intelligence, {IJCAI-20}}, pages
  3349--3355. International Joint Conferences on Artificial Intelligence
  Organization, 7 2020.
\newblock \doi{10.24963/ijcai.2020/463}.
\newblock URL \url{https://doi.org/10.24963/ijcai.2020/463}.
\newblock Main track.

\bibitem[Singh(1992)]{Singh1992}
Satinder~Pal Singh.
\newblock Transfer of learning by composing solutions of elemental sequential
  tasks.
\newblock \emph{Mach. Learn.}, 8\penalty0 (3):\penalty0 323--339, May 1992.

\bibitem[Sinha and Garg(2021)]{sinha2021s4rl}
Samarth Sinha and Animesh Garg.
\newblock S4rl: Surprisingly simple self-supervision for offline reinforcement
  learning.
\newblock \emph{arXiv preprint arXiv:2103.06326}, 2021.

\bibitem[Sohn et~al.(2015)Sohn, Lee, and Yan]{sohn2015-vae}
Kihyuk Sohn, Honglak Lee, and Xinchen Yan.
\newblock Learning structured output representation using deep conditional
  generative models.
\newblock In C.~Cortes, N.~Lawrence, D.~Lee, M.~Sugiyama, and R.~Garnett,
  editors, \emph{Advances in Neural Information Processing Systems}, volume~28.
  Curran Associates, Inc., 2015.
\newblock URL
  \url{https://proceedings.neurips.cc/paper/2015/file/8d55a249e6baa5c06772297520da2051-Paper.pdf}.

\bibitem[Spall et~al.(2012)Spall, Wallentin, Yusuf, Eikelboom, Nieuwlaat, Yang,
  Kabali, Reilly, Ezekowitz, Connolly, and et~al.]{van_spall}
Harriette G.c.~Van Spall, Lars Wallentin, Salim Yusuf, John~W. Eikelboom, Robby
  Nieuwlaat, Sean Yang, Conrad Kabali, Paul~A. Reilly, Michael~D. Ezekowitz,
  Stuart~J. Connolly, and et~al.
\newblock Variation in warfarin dose adjustment practice is responsible for
  differences in the quality of anticoagulation control between centers and
  countries.
\newblock \emph{Circulation}, 126\penalty0 (19):\penalty0 2309–2316, 2012.
\newblock \doi{10.1161/circulationaha.112.101808}.

\bibitem[Suresh et~al.(2017)Suresh, Hunt, Johnson, Celi, Szolovits, and
  Ghassemi]{suresh2017clinical}
Harini Suresh, Nathan Hunt, Alistair Johnson, Leo~Anthony Celi, Peter
  Szolovits, and Marzyeh Ghassemi.
\newblock Clinical intervention prediction and understanding using deep
  networks.
\newblock \emph{JMLR (Journal of Machine Learning Research): MLHC Conference
  Proceedings}, 2017.

\bibitem[Sutton and Barto(2018)]{sutton_book}
R.~S. Sutton and A.~G. Barto.
\newblock \emph{Reinforcement Learning: An Introduction}.
\newblock MIT Press, 2nd edition, 2018.
\newblock ISBN 9780262039246.

\bibitem[Sutton(1995)]{Sutton1995modeling}
Richard~S. Sutton.
\newblock {TD} models: Modeling the world at a mixture of time scales.
\newblock In Armand Prieditis and Stuart Russell, editors, \emph{Machine
  Learning Proceedings 1995}, pages 531--539. Morgan Kaufmann, San Francisco
  (CA), 1995.
\newblock ISBN 978-1-55860-377-6.
\newblock \doi{https://doi.org/10.1016/B978-1-55860-377-6.50072-4}.

\bibitem[Sutton et~al.(1999)Sutton, Precup, and Singh]{sutton-smdp}
Richard~S. Sutton, Doina Precup, and Satinder Singh.
\newblock Between mdps and semi-mdps: A framework for temporal abstraction in
  reinforcement learning.
\newblock \emph{Artificial Intelligence}, 112\penalty0 (1):\penalty0 181--211,
  1999.
\newblock ISSN 0004-3702.

\bibitem[Tang et~al.(2020{\natexlab{a}})Tang, Modi, Sjoding, and
  Wiens]{shengpu2020}
Shengpu Tang, Aditya Modi, Michael Sjoding, and Jenna Wiens.
\newblock {Clinician-in-the-Loop} decision making: Reinforcement learning with
  {Near-Optimal} {Set-Valued} policies.
\newblock 119:\penalty0 9387--9396, 2020{\natexlab{a}}.

\bibitem[Tang et~al.(2020{\natexlab{b}})Tang, Modi, Sjoding, and
  Wiens]{tang2020clinician}
Shengpu Tang, Aditya Modi, Michael Sjoding, and Jenna Wiens.
\newblock Clinician-in-the-loop decision making: Reinforcement learning with
  near-optimal set-valued policies.
\newblock In \emph{International Conference on Machine Learning}, pages
  9387--9396. PMLR, 2020{\natexlab{b}}.

\bibitem[Watkins(1989)]{Watkins89-phd}
C.~J. C.~H. Watkins.
\newblock \emph{Learning from Delayed Rewards}.
\newblock PhD thesis, King's College, 1989.

\bibitem[Watkins and Dayan(1992)]{Watkins1992-ty}
Christopher J C~H Watkins and Peter Dayan.
\newblock Q-learning.
\newblock \emph{Mach. Learn.}, 8\penalty0 (3):\penalty0 279--292, May 1992.

\bibitem[(WHO)(2021)]{who}
World Health~Organization (WHO).
\newblock \emph{World Health Statistics 2021: monitoring health for the SDGs,
  sustainable development goals}.
\newblock Licence: CC BY-NC-SA 3.0 IGO., 2021.

\bibitem[Wixson(1991)]{WIXSON1991368}
Lambert~E. Wixson.
\newblock Scaling reinforcement learning techniques via modularity.
\newblock In Lawrence~A. Birnbaum and Gregg~C. Collins, editors, \emph{Machine
  Learning Proceedings 1991}, pages 368--372. Morgan Kaufmann, San Francisco
  (CA), 1991.
\newblock ISBN 978-1-55860-200-7.
\newblock \doi{https://doi.org/10.1016/B978-1-55860-200-7.50076-3}.

\bibitem[Wu et~al.(2016)Wu, Ghassemi, Feng, Celi, Szolovits, and
  Doshi-Velez]{wu2016ssam}
Mike Wu, Marzyeh Ghassemi, Mengling Feng, Leo~A Celi, Peter Szolovits, and
  Finale Doshi-Velez.
\newblock Understanding vasopressor intervention and weaning: Risk prediction
  in a public heterogeneous clinical time series database.
\newblock \emph{Journal of the American Medical Informatics Association}, page
  ocw138, 2016.

\bibitem[Yu et~al.(2019)Yu, Liu, and Nemati]{yu2019reinforcement}
Chao Yu, Jiming Liu, and Shamim Nemati.
\newblock Reinforcement learning in healthcare: a survey.
\newblock \emph{arXiv preprint arXiv:1908.08796}, 2019.

\bibitem[Zarezade et~al.(2017)Zarezade, De, Upadhyay, Rabiee, and
  Gomez-Rodriguez]{Zarezade2017}
Ali Zarezade, Abir De, Utkarsh Upadhyay, Hamid~R. Rabiee, and Manuel
  Gomez-Rodriguez.
\newblock Steering social activity: A stochastic optimal control point of view.
\newblock \emph{J. Mach. Learn. Res.}, 18\penalty0 (1):\penalty0 7512–7546,
  jan 2017.
\newblock ISSN 1532-4435.

\end{thebibliography}

\clearpage

\appendix

\section{Illustrative Example}\label{apd:toy_example_appendix}

\begin{figure}[b!]
    \centering
    \includegraphics[width=0.4\linewidth]{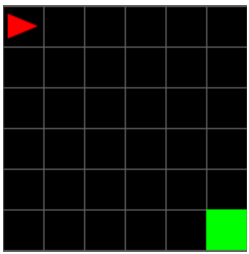}
    \caption{Minigrid 8x8 environment.}
    \label{fig:minigrid:appdx}
\end{figure}

This section will document additional details of the illustrative example. We use the Gym Minigrid environment \citep{gym_minigrid} with an 8x8 grid, depicted in Figure \ref{fig:minigrid:appdx}. We use additional wrappers to modify the state space, observations, and reward structure. Most notably, we introduce options as discussed in the main body of the paper. We will release the code after the review process is concluded.

\subsection{Model Architecture}
We use the same fully-connected network architecture for all three algorithms: Semi Deep Q-Netowrk (SDDQN), Semi Double DQN (SDDQN), and Semi Batch-Constrained Learning (SBCQ). The state space, after converting to the flat array, has a dimension size of 108. The fully-connected network has two hidden layers, with 128 nodes and 64 nodes, followed by a linear layer which outputs the $Q$ value of each action/option. The ReLU activation function is applied to the two hidden layers. The networks are trained using a learning rate of $0.0005$ and minibatch size of 32. For the SBCQ algorithm, we use a BCQ threshold of 0.3 for all of the experiments. Other global hyper parameters can be found in the \texttt{config.yaml} file at the root directory of the code.

\subsection{Online Experiment Details}
There are two reward signals in the environment: +10 when the agent reaches the goal state, and -1 when the agent crosses the third row (unless the agent is in the rightmost column). There are three possible options to take: turn left, turn right, or move forward. Each option consists of the respective primitive action, followed by a sequence of forward steps. The step size varies by location -- when the agent is on the top row, the step size is 1. When the agent is in the leftmost column facing downwards, the step size is 4. Elsewhere, the step size is 2. The discount factor in the online setting is 0.9. 

To ensure adequate exploration during training, we begin with an epsilon of 1.0 (full exploration) for 5,000 training steps, and we anneal the epsilon to 0.1 over 10,000 additional steps. After every 20 training episodes, the current policy is evaluated in the environment by running one episode with the agent starting at the default initial position (upper left corner). When the final policy is animated, we observe that the MDP-based algorithms choose the path that incurs the negative reward, whereas the SMDP-based algorithms avoid the negative reward by taking smaller steps towards the goal.

\subsection{Offline Experiment Details}
The results in the offline experiment are run on a slightly modified environment. The first difference is that the agent incurs a reward of -3 instead of -1 for crossing the third row. The second difference is that there are only two step sizes: a step size of 4 when the agent is in the leftmost column facing downwards, and a step size of 2 elsewhere. The third difference is that the discount factor used in the offline setting is 0.95 instead of 0.9.

\begin{figure*}[h!]
    \centering
    \includegraphics{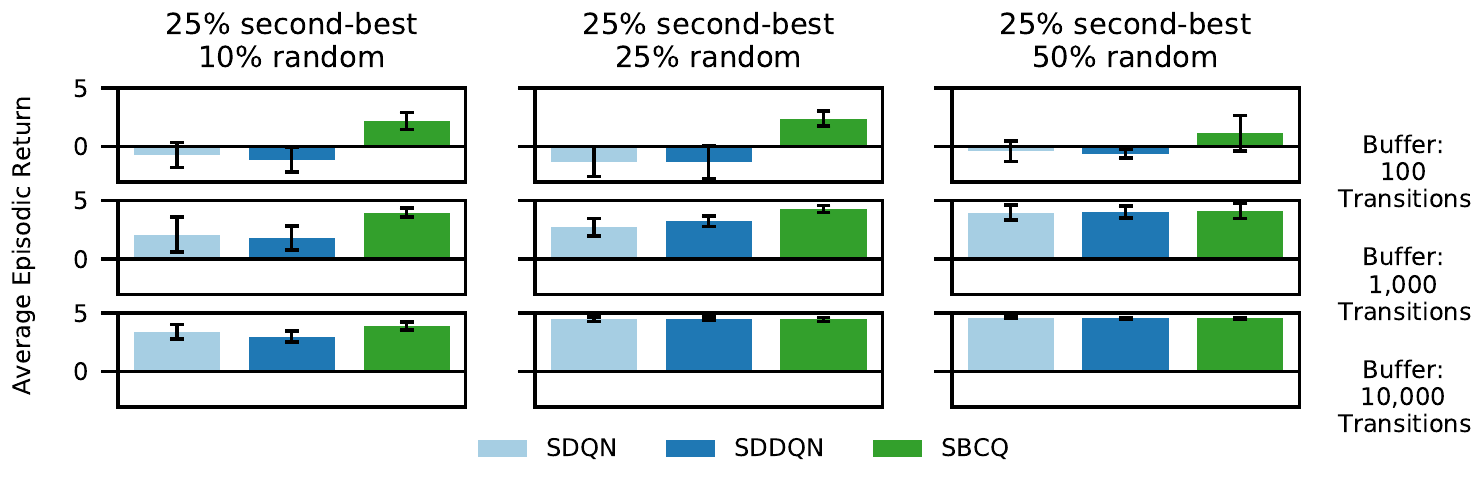}
    \caption{Episodic returns averaged across all test cases for each dataset configuration and each of SDQN, SDDQN, and SBCQ. Model selection was done using a fixed validation dataset (with 25\% random decision and 25\% second-best decisions) across all experiments.}
    \label{fig:buffer_fixed_val}
\end{figure*}

\begin{figure*}[h!]
    \centering
    \includegraphics{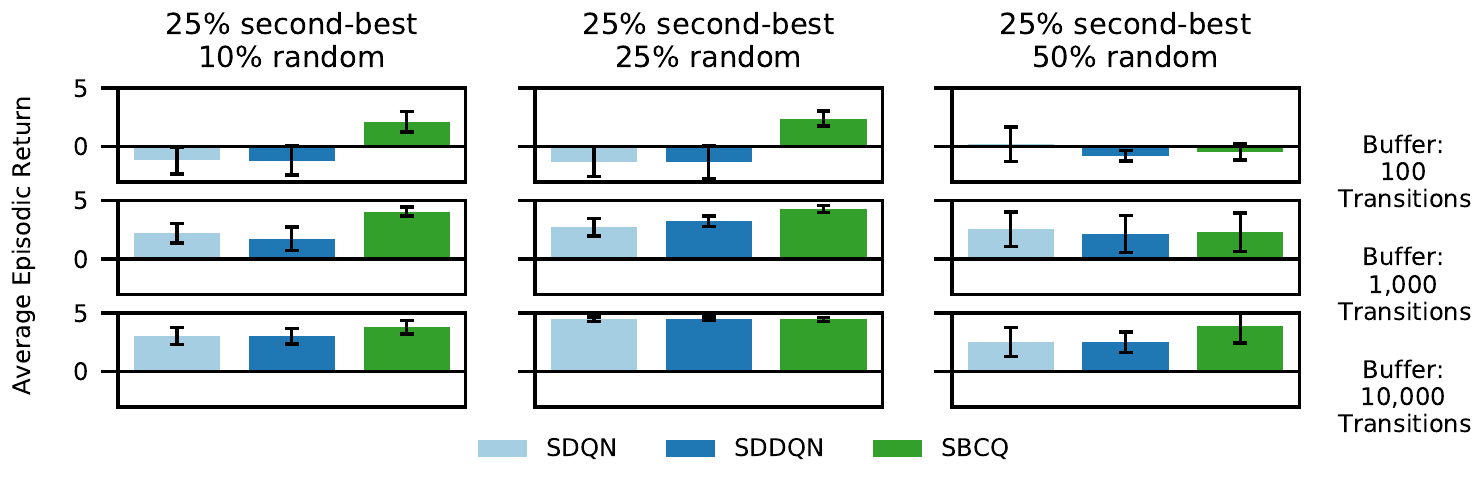}
    \caption{Episodic returns averaged across all test cases for each dataset configuration and each of SDQN, SDDQN, and SBCQ. Model selection was done using a validation dataset with the same composition as the training dataset.}
    \label{fig:buffer_analysis_same_comp}
\end{figure*}

To have a fully offline learning scenario, the validation dataset is offline as well. There are a few ways to select the validation data; e.g., using the same composition as the training data, or using the same validation dataset across all experiments. Regardless of the selected approach, the validation dataset is composed of 250 transitions and is used to evaluate the network under training in order to choose the best training epoch. The best network with the lowest validation loss is then selected to evaluate on the test data. In the results presented in the main text, the models were selected using the second validation approach. Figures \ref{fig:buffer_fixed_val} and  \ref{fig:buffer_analysis_same_comp} show the results using the two validation approaches. The results presented in the figures are aggregated across nine random seeds. In the final version, we will update the main text to reflect the results across more random seeds.

\newpage
\section{Warfarin Dosing}\label{apd:warfarin_appendix}

Access to the warfarin data used for this study is governed by an executive committee. Details regarding the data and how to reach the committee for potential collaborations will be released after the review process is concluded.

\subsection{Data Preprocessing}
The warfarin dataset begins with 29,272 adult patients across the four trials. We begin by excluding patients who did not have warfarin and INR data, as well as patients who had weekly doses exceeding 140mg. This leaves us with 28,444 patients. We also remove entries that were recorded prior to the first INR measurement and after the final INR measurement. 

For the state space, we extract the features deemed relevant based on clinical input. These features were age, sex, weight, region, smoking status, concurrent medications (aspirin, amiodarone, thienopyridines), and medical history (of diabetes, hypertension, myocardial infarction). The state space also consists of the current INR, the previous four INR measurements, the previous warfarin dose, and binary variables indicating whether the patient experienced ischemic stroke, hemorrhagic stroke, major bleeding, or hospitalizations during the treatment. All of the continuous features are normalized, and the categorical features are converted to one-hot encodings. To improve training, the continuous features are also discretized by one-hot encoding the values into quantiles (hence, both normalized and quantile versions are used). The fixed, demographic features and the time-varying features are summarized in Tables \ref{tab:warfarin_stats_demog} and \ref{tab:warfarin_stats_time}, respectively. The final state space has a dimension size of 56.

\begin{table*}[h]
\caption{Warfarin Demographic Features} \label{tab:warfarin_stats_demog}
\begin{center}
\begin{tabular}{ll}
\textbf{Feature} &\textbf{Data Format} \\
\hline \\ 
Sex & Binary Encoding \\
Weight & Float (Min-Max Scaling) \\
Weight (Discretized) & One-Hot Encoding (6 Quantile Categories)  \\
Age & Float (Min-Max Scaling)  \\
Age (Discretized) & One-Hot Encoding (7 Quantile Categories)  \\
Continent & One-Hot Encoding (6 Categories) \\
On Aspirin & Binary Encoding \\
On Amiodarone & Binary Encoding \\
On Thienopyridine & Binary Encoding \\
History of Diabetes & Binary Encoding \\
History of Myocardial Infarction & Binary Encoding \\
History of Congestive Heart Failure  & Binary Encoding \\
History of Hypertension  & Binary Encoding \\
History of Smoking & One-Hot Encoding (3 Categories) \\

\end{tabular}
\end{center}
\end{table*}

\begin{table*}[h]
\caption{Warfarin Time-Varying Features} \label{tab:warfarin_stats_time}
\begin{center}
\begin{tabular}{ll}
\textbf{Feature} &\textbf{Data Format} \\
\hline \\ 
Previous Warfarin Dose & Float (Min-Max Scaling) \\
Previous Warfarin Dose (Discretized) & One-Hot Encoding (10 Quantile Categories) \\
Current INR & Float (Min-Max Scaling) \\
Current INR (Discretized) & One-Hot Encoding (5 Quantile Categories) \\
Previous Four INR Measurements & Float (Min-Max Scaling) \\

Minor Bleed Occurrence & Binary Encoding \\
Major Bleed Occurrence & Binary Encoding \\
Hospitalization Occurrence & Binary Encoding \\

\end{tabular}
\end{center}
\end{table*}

To construct the options, we use the recorded clinical decisions and discretized the dose increments based on clinical expertise and existing clinical algorithms \citep{van_spall, Nieuwlaat2012RandomisedCO}. Following clinical practice, the dose increments used in the underlying primitive actions are: maintain current dosage, increase in increments of $<$10\%, 10-20\%, and over 20\%, or decrease in the same increments. This yields seven primitive actions. Options are then constructed as one of the seven primitive actions, followed by the primitive action of maintaining the current dosage. 

There were a few cases where a patient treatment was split into multiple trajectories. The first case is when adverse medical events occur during the treatment, and the records surrounding the events may not be as accurate as other records. As such, the trajectory ends when an adverse event occurs, and the remaining treatment is treated as a new trajectory. The second case is when more than 90 days elapsed between clinical visits. Since this is a longer time elapsed than clinical recommendation \citep{van_spall}, it is deemed that interpolation between these points (for reward estimation) is unreliable. Instead, we split this trajectory into two trajectories. After creating trajectories from the patient data, some trajectories were very short (the shortest trajectory was only one measurement). As a result, we also remove trajectories that have fewer than ten decisions.

To split the data, we split along the trials. Half of the ARISTOTLE patients were held out for testing, and the remainder of the ARISTOTLE was split between training and validation. The validation data was generated by randomly sampling from the remaining ARISTOTLE data, as well as the RE-LY data. The rest of the data was used as the training data.

\subsection{Model Architectures}
We use the same fully-connected network for both the DDQN and BCQ. The state space, as described in the previous section, has a dimension size of 56. The fully-connected network has two hidden layers, with 64 nodes in each, followed by a linear layer mapping to the $Q$ value of each option. The ReLU activation function is applied to both hidden layers. The networks are trained using the learning rate of $5 \times 10^{-5}$ and batch size of 64. For the SBCQ algorithm, the BCQ threshold is 0.2.

\subsection{Model Validation}
To select the model to use on the test data, we estimate what the time in therapeutic range (TTR) would be had the clinician followed the policy. This is estimated by evaluating the trajectories where there is good agreement between the policy and the actions chosen by the clinician in the same setting. Each of the seven options corresponds an integer as such: [0: Decrease dose by $>20\%$, 1:  Decrease dose by 10-20\%, 2:  Decrease dose by 10-20\%, 3: Maintain current dose, 4: Increase dose by $<10\%$, 5: Increase dose by $10-20\%$, 6: Increase dose by $>20\%$]. The larger the difference between the doses, the larger the difference in the clinical outcome of the dose decision. In other words, we denote the difference between the policy and clinician decision as the absolute difference between the corresponding integers of their respective decisions. We can then sum up the differences along a trajectory, to estimate the agreement between the policy and clinician for any given trajectory. Then, we average the observed TTR for trajectories whose agreement exceeds 85\% of the trajectories. This yields the estimated TTR for following the policy. 

\end{document}